\documentclass[11pt]{article}
\usepackage{iclr2026_conference,times}
\usepackage[margin=1in]{geometry}
\usepackage{amsmath,amssymb}
\usepackage{booktabs}
\usepackage{multirow}
\usepackage{makecell}
\usepackage{tabularx}
\usepackage{graphicx}
\usepackage{subcaption}
\usepackage{xcolor}
\usepackage{hyperref}
\usepackage[nameinlink,capitalise]{cleveref}
\usepackage{natbib}
\usepackage{url}
\usepackage{fancyvrb}
\usepackage{fvextra}
\usepackage{multicol}
\usepackage{enumitem}
\usepackage{xurl}
\usepackage[normalem]{ulem}

\usepackage[most]{tcolorbox}
\usepackage{listings}
\usepackage[table]{xcolor}

\newcommand{\op}[1]{\item \path{#1}}
\definecolor{crimson}{HTML}{DC143C}

\iclrfinalcopy

\definecolor{mtblue}{HTML}{C2410C}

\newcommand{\makeMusaTitle}{
\begin{center}

\vspace*{-6.3em}

\begin{minipage}{\textwidth}
    \raggedright
    \includegraphics[height=0.65cm]{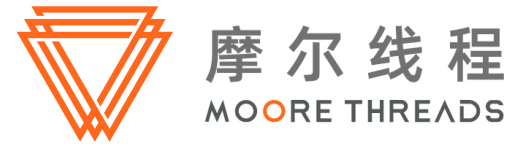}
\end{minipage}

\rule{\textwidth}{0.8pt}

{\Large\bfseries TCPO: Turn-Level Credit Policy Optimization}

\vspace{0.6em}

{\large\bfseries
Sicong Liao,\quad
Zhi Chen,\quad
Yaohua Tang
}

\vspace{0.1em}

{\normalsize
Moore Threads AI\\[0.25em]
\texttt{tangyaohua28@gmail.com}
}

\vspace{0.8em}

\end{center}
}

\begin{document}
\makeMusaTitle

\begin{abstract}
Verifier-guided reinforcement learning has become a powerful paradigm for improving LLM reasoning. In multi-turn settings, models receive a verifier score after each turn and iteratively refine their outputs. Although such scores provide dense feedback, they do not directly provide dense credit: a score measures the quality of the current output, while credit should measure how the current turn changes the refinement trajectory. We propose TCPO, a turn-level credit assignment method for verifier-guided multi-turn RL. TCPO casts credit assignment as \emph{score-to-credit conversion} and constructs turn-level advantages through reference-based comparisons: retrospective credit captures immediate progress and regression relative to the best prior state; hindsight delayed credit identifies non-improving turns with later payoff; and selective fixed-history counterfactual estimation refines high-surprisal turns under the same history.  Experiments on math reasoning, code generation, and AppWorld agent tasks show that TCPO improves or matches the strongest baselines across model scales, task domains, and verifier types. TCPO achieves the best or tied-best best-turn Pass@8 on Qwen3-4B and DeepSeek-R1-Distill-Llama-8B, reduces turns to success, and improves multi-turn agent performance. These results highlight score-to-credit conversion as a central ingredient for verifier-guided multi-turn policy optimization.
\end{abstract}

\section{Introduction}

Reinforcement learning with verifiable rewards (RLVR) has become a central paradigm for improving the mathematical reasoning, code generation, and tool-use abilities of large language models. As LLM agents move from single-turn generation to code execution, tool invocation, search, and interactive refinement, optimization increasingly takes the form of verifier-guided multi-turn RL: at each turn, the policy generates a candidate output, receives a verifier score, and revises its response until success or a turn limit is reached. Compared with outcome-only supervision, this setting provides richer process feedback.

Yet denser verifier scores do not automatically yield better credit assignment. A verifier score measures the quality of the current output, whereas turn-level credit should measure the contribution of the current turn to the overall refinement process. A high-scoring turn may simply preserve an already repaired state; a non-improving turn may enable later success; and a post-success modification may corrupt a correct answer. Thus, the central challenge is not merely obtaining per-turn signals, but converting existing verifier scores into optimizable turn-level credit. We call this problem \emph{score-to-credit conversion}.

Existing approaches do not explicitly solve this conversion problem. Trajectory-level methods broadcast final or best-turn success to all tokens, mixing failed attempts, useful repairs, and post-success regressions. Raw turn score rewards state quality rather than marginal contribution. Adjacent score difference captures local change, but cannot tell whether a turn improves upon the best-so-far state or whether a non-improving turn later proves useful. Although recent turn-level reward design and agentic RL methods highlight the mismatch between multi-turn decisions and trajectory-level supervision, how to assign already-observed verifier scores to individual turns remains underexplored.

We propose TCPO (\emph{Turn-Level Credit Policy Optimization}), a credit-assignment method for verifier-guided multi-turn RL. TCPO leaves the base multi-turn rollout unchanged and requires no additional human labels or learned critic. After base rollouts are completed, it converts verifier traces into turn-level advantages through three reference comparisons. \emph{Retrospective credit} compares each turn with the best score achieved before it, capturing immediate progress, success preservation, and regression. For eligible pre-success non-improving turns, \emph{hindsight delayed credit} compares their realized future-best score against same-prompt, same-turn eligible trajectories. For high-surprisal eligible turns, \emph{fixed-history counterfactual estimation} samples alternative outputs under the same history and compares them with the original output using the verifier. The resulting turn-level advantages are broadcast to the corresponding response tokens and can be used with GRPO-style objectives.

Empirically, TCPO is effective across model scales, task domains, and verifier types. On math and code benchmarks, TCPO achieves the best or tied-best best-turn Pass@8 on both Qwen3-4B and DeepSeek-R1-Distill-Llama-8B, while usually reducing turns to success. Compared with MT-GRPO, TCPO improves Qwen3-4B by $+4.4$, $+2.2$, $+1.6$, and $+0.6$ points on MATH-500, AIME, LiveCodeBench, and HumanEval, respectively, with even larger gains on DeepSeek-R1-Distill-Llama-8B. TCPO also outperforms or matches GVPO. On AppWorld, TCPO improves Dev and Test-N on both TGC and SGC and improves Test-C on TGC, demonstrating transfer to multi-turn tool-using agents.

Our contributions are:
\begin{itemize}
    \item We formulate \emph{score-to-credit conversion}: in verifier-guided multi-turn RL, the key challenge is not reward density but assigning per-turn verifier scores as marginal turn-level credit.
    \item We introduce TCPO, which constructs turn-level advantages using retrospective, hindsight, and fixed-history counterfactual references to address immediate progress, delayed contribution, and high-surprisal credit ambiguity.
    \item We validate TCPO on math, code, and AppWorld agent tasks, showing improvements over or parity with strong baselines at modest additional training cost.
\end{itemize}

\section{Related Work}

Reinforcement learning has increasingly been used to train LLM agents across a wide range of interactive settings, including code execution and computational tool use~\citep{feng2025retool,li2025torl}, general tool use and API-based environments~\citep{wang2025otc,zhang2025nemotron,singh2025agentic,chen2025reinforcement}, reasoning with search engines~\citep{chen2026learning,jin2025search,jin2025empirical}, web navigation and graphical-interface control~\citep{yao2022webshop,bai2024digirl}, and games, embodied tasks, and other interactive environments~\citep{yao2020keep,carta2023grounding,zhai2024fine,wang2025ragen}. 
Unlike single-turn reasoning, these settings require the model to make a sequence of decisions, including generating intermediate solutions, interpreting feedback, revising previous outputs, invoking tools, and deciding when to stop. 
However, many agent RL pipelines still reduce the interaction to a trajectory-level return. 
This creates a granularity mismatch between decision and supervision: the agent acts turn by turn, but the reward does not identify which turn repaired the solution, preserved a solved state, introduced redundant changes, or caused a regression. 
Recent work therefore argues that multi-turn optimization should preserve this sequential, turn-level decision structure rather than collapsing the entire interaction into a single trajectory-level optimization problem~\citep{zeng2025reinforcing,li2026turn,tao2026trace,zong20262}.

Prior work addresses this mismatch through finer-grained supervision or advantage estimation. 
PPO provides token-level credit through a learned value model, but critic training adds complexity and can be brittle on long reasoning chains~\citep{zhou2025sweet,chen2025context,tao2026trace}. 
Critic-free methods such as GRPO avoid value modeling by using group-relative sequence rewards, but their sequence-level advantages remain coarse for long or multi-turn trajectories~\citep{shao2024deepseekmath}. 
Tree- or prefix-based methods exploit branching structures among sampled responses to derive denser token- or prefix-level credit without training a separate critic~\citep{tran2026exploiting,zong20262}. 
In parallel, process-supervision and step-level reward methods use intermediate labels, process reward models, learned evaluators, or process-verifiable feedback to provide denser supervision~\citep{uesato2022solving,lightman2024let,wang2024math,dai2026group}. 
These approaches establish the importance of fine-grained supervision, but they mainly address how to obtain denser signals, propagate sparse outcome rewards, or align optimization with turn-level decisions. 
TCPO addresses a complementary problem: when verifier scores are already available at every turn, how should these scores be converted into turn-level credit? 
It separates immediate score changes from delayed contributions to subsequent repair and selectively refines ambiguous turns through fixed-history counterfactual comparisons. 
Thus, TCPO is a credit-assignment method for verifier-guided multi-turn RL, rather than an additional source of intermediate rewards.

\begin{figure*}[t]
\centering
\includegraphics[width=\textwidth]{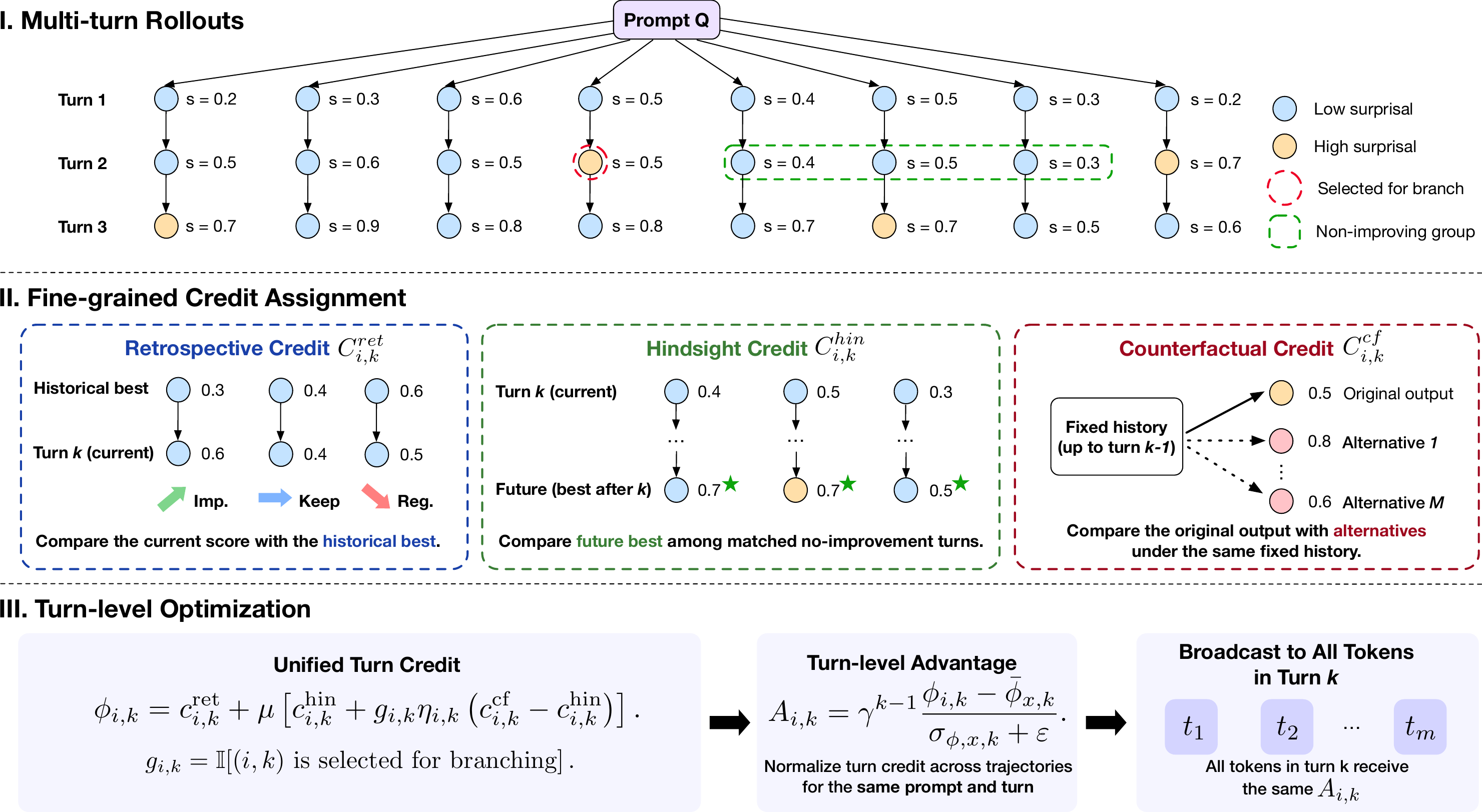}
\caption{Overview of TCPO. Given a prompt, the policy samples multiple verifier-guided multi-turn trajectories. TCPO converts per-turn verifier scores into structured turn-level credit by combining progress-to-go rewards, keep rewards, regression penalties, budgeted fixed-history counterfactual credit, and early-turn weighting. The resulting turn-level advantages are broadcast to response tokens and optimized with a GRPO-style objective.}
\label{fig:TCPO-overview}
\end{figure*}

\section{Method}

Figure~\ref{fig:TCPO-overview} summarizes the TCPO pipeline. For each prompt, we first sample a group of multi-turn trajectories and record verifier scores at every turn. TCPO then converts these score traces into turn-level credit through three reference comparisons: retrospective credit against the best previous state, hindsight delayed credit from the future-best quality in the completed rollout, and budgeted fixed-history counterfactual refinement for high-surprisal pre-success non-improving turns. The resulting turn credit is normalized within prompt-turn groups, reweighted by an early-turn prior, broadcast to the tokens of the corresponding response, and optimized with a GRPO-style clipped objective.

\paragraph{Multi-Turn Verifier Setting}

Given a prompt $x$, the policy interacts with a verifier over multiple turns. 
At each turn, the model generates a candidate output, receives verifier feedback, and revises its solution for a bounded number of turns. 
For the $k$-th turn of the $i$-th trajectory, let the interaction state be
\[
s_{i,k}=(x,T_{i,<k},F_{i,<k},E_{i,k}),
\]
where $T_{i,<k}$ is the transcript of previous model outputs, tool calls, and observations, $F_{i,<k}$ contains verifier scores and textual feedback, and $E_{i,k}$ is the external environment state before turn $k$ when the task is stateful. For math and code tasks, $E_{i,k}$ is empty; for AppWorld, it is the simulator state snapshot before the next action. We write $h_{i,k}$ as a shorthand for this full state in the equations below. 
The rollout policy samples the current output, and the verifier returns a score:
\[
y_{i,k}\sim \pi_{\mathrm{old}}(\cdot\mid h_{i,k}),
\qquad
r_{i,k}=\mathcal{V}(x,h_{i,k},y_{i,k}).
\]
Here, $\mathcal{V}$ is a non-differentiable verifier such as an answer checker or a code test suite, and all verifier scores are normalized so that $r_{i,k}\in[0,1]$. 
During training, rollouts are generated for a fixed maximum budget $K$ even if a trajectory first reaches the success threshold before turn $K$; the first-success turn is recorded but later turns are still scored. This fixed-budget training protocol makes preservation and regression after a solved state observable. During evaluation, success and average turns are computed from the first successful turn. For math tasks, success means producing the correct answer, and for code tasks, success means passing all test cases. 
Let $\mathcal{A}_{x,k}$ denote the set of trajectories for prompt $x$ that contain a response at turn $k$.

\paragraph{Reference-Aware Credit Decomposition}

Verifier scores provide turn-by-turn observations of output quality, but they are not themselves turn-level credit. 
TCPO decomposes credit according to three references: the best score achieved before the turn, the realized future after the turn, and alternative outputs under the same history. 
The next three subsubsections instantiate these references as retrospective credit, hindsight estimation, and fixed-history counterfactual estimation.

\subsubsection{Retrospective Credit}

Let
\[
m_{i,k}=\max_{t<k}r_{i,t},
\qquad
\Delta_{i,k}=[r_{i,k}-m_{i,k}]_+,
\]
where $m_{i,1}=0$. 
The retrospective credit is defined as
\[
\begin{aligned}
c^{\mathrm{ret}}_{i,k}
&=
\alpha_+\Delta_{i,k}
+\alpha_0 I^{\mathrm{keep}}_{i,k}
-\alpha_- I^{\mathrm{succ}}_{i,k}[m_{i,k}-r_{i,k}]_+,\\
I^{\mathrm{succ}}_{i,k}
&=
\mathbf{1}[m_{i,k}\ge\tau],\\
I^{\mathrm{keep}}_{i,k}
&=
I^{\mathrm{succ}}_{i,k}
\mathbf{1}[\Delta_{i,k}=0]\mathbf{1}[r_{i,k}\ge m_{i,k}].
\end{aligned}
\]
The first term rewards a turn that improves upon the best score achieved before the turn. The second term rewards preservation: once the trajectory has already reached a successful state, the current turn receives credit if it keeps the solution at the best-so-far score without further improvement. The third term penalizes regression after success, i.e., cases where the trajectory has previously reached the success threshold but the current turn lowers the verifier score. The preservation term is active only when $\Delta_{i,k}=0$, making it mutually exclusive with progress. This signal is defined for all turns, but for turns with $m_{i,k}<\tau$ and $\Delta_{i,k}=0$, it is identically zero. Such turns cannot be distinguished from immediate score changes alone and require a future-based reference.

\paragraph{Hindsight Estimation}

We call a turn \emph{non-improving} if it does not exceed the best-so-far verifier score, i.e., $\Delta_{i,k}=0$. Delayed credit is assigned only to pre-success non-improving turns, because post-success preservation and regression are already handled by retrospective credit. We define the eligibility gate
\[
b_{i,k}
=
\mathbf{1}[m_{i,k}<\tau]\mathbf{1}[\Delta_{i,k}=0].
\]
Define the future-best score after turn $k$ as
\[
u_{i,k}=\max_{t\ge k}r_{i,t}.
\]
The reference set consists of trajectories under the same prompt and turn index that satisfy the same eligibility gate. 
We use a leave-one-out baseline:
\[
\bar u^{-i}_{x,k}
=
\operatorname{Avg}_{\substack{
j\in\mathcal{A}_{x,k},\,j\ne i\\
b_{j,k}=1}}
u_{j,k},
\]
with a fallback of zero hindsight credit if the set is empty. 
The hindsight estimate is
\[
c^{\mathrm{hin}}_{i,k}
=
b_{i,k}\cdot
\big(u_{i,k}-\bar u^{-i}_{x,k}\big).
\]
The gate prevents overlap with retrospective progress, keep, and regression terms. 
The sign indicates whether an eligible non-improving turn led to better or worse future-best quality than comparable eligible turns. 
This estimate uses only existing rollout information and requires no additional verifier calls.

\paragraph{Counterfactual Estimation and Budgeted Refinement}

The hindsight estimate compares realized futures across different trajectories, rather than alternative choices under the same history. 
It may therefore attribute later corrections to the current turn. 
To refine the most ambiguous cases, TCPO performs fixed-history counterfactual estimation for high-surprisal eligible non-improving turns.

We use average surprisal of the sampled response as a proxy for credit uncertainty:
\[
\xi_{i,k}
=
-\frac{1}{|y_{i,k}|}
\sum_t
\log \pi_{\mathrm{old}}(a_{i,k,t}\mid h_{i,k},a_{i,k,<t}),
\]
and select up to $L$ eligible non-improving turns for each prompt:
\[
\begin{aligned}
\mathcal{B}_{x}
&=
\{(i,k): i\in\mathcal{A}_{x,k},\, b_{i,k}=1\},\\
\mathcal{I}_{x}
&=
\operatorname{TopL}_{(i,k)\in\mathcal{B}_{x}}(\xi_{i,k}).
\end{aligned}
\]
where early-turn weighting is applied later through the turn advantage.

For each selected turn $(i,k)\in\mathcal{I}_{x}$, we fix $h_{i,k}$ and sample $M$ alternative outputs from the rollout policy:
\[
\tilde y^{(m)}_{i,k}\sim \pi_{\mathrm{old}}(\cdot\mid h_{i,k}),
\quad m=1,\ldots,M,
\qquad
\tilde y^{(0)}_{i,k}=y_{i,k}.
\]
For stateful environments, all alternatives are evaluated from the same pre-turn environment snapshot $E_{i,k}$ and the simulator is restored after each alternative, so the transcript, feedback, and environment state are fixed across the comparison. 
Each candidate is scored by the verifier:
\[
q_m=\mathcal{V}(x,h_{i,k},\tilde y^{(m)}_{i,k}).
\]
The counterfactual estimate is
\[
c^{\mathrm{cf}}_{i,k}
=
q_0-\frac{1}{M}\sum_{m=1}^{M}q_m.
\]

We use one-step verifier comparison rather than full counterfactual rollouts to isolate the effect of the current output and avoid mixing in later revisions. This choice is conservative for binary exact-match verifiers: if the original output and all alternatives fail, the counterfactual signal is zero, and if an alternative succeeds while the original fails, the original receives negative local credit. Positive credit for latent useful repairs in such cases is therefore supplied by the hindsight term, while the counterfactual term acts as a budgeted local refinement rather than a standalone delayed-credit oracle.

The delayed credit is assigned by a non-degeneracy-gated refinement of the low-cost hindsight estimate. Let
\[
g_{i,k}=\mathbf{1}[(i,k)\in\mathcal{I}_{x}],
\qquad
\eta_{i,k}=\mathbf{1}[\operatorname{Var}(q_0,\ldots,q_M)>0],
\]
where $\eta_{i,k}$ indicates whether the fixed-history alternatives produce a non-degenerate verifier comparison. We define
\[
\begin{aligned}
c^{\mathrm{del}}_{i,k}
&=
c^{\mathrm{hin}}_{i,k}
+g_{i,k}\eta_{i,k}
\left(
c^{\mathrm{cf}}_{i,k}-c^{\mathrm{hin}}_{i,k}
\right).
\end{aligned}
\]
Thus, selected turns use the fixed-history comparison only when the alternatives reveal score variation; otherwise TCPO falls back to hindsight credit. This avoids deleting positive hindsight credit for latent useful repairs when binary alternatives all receive the same verifier score, while still allowing informative counterfactual comparisons to override noisier cross-trajectory hindsight estimates.

\paragraph{Turn-Level Advantage and Token Optimization}

TCPO combines retrospective and delayed credit into a raw turn signal:
\[
\phi_{i,k}
=
c^{\mathrm{ret}}_{i,k}
+
\mu c^{\mathrm{del}}_{i,k}.
\]
Since both hindsight estimation and counterfactual selection are restricted to $b_{i,k}=1$, delayed credit is zero for turns with immediate improvement and for post-success keep or regression turns. These cases are handled by retrospective credit.

We normalize the raw signal within the same prompt and turn, then apply an early-turn prior:
\[
\begin{aligned}
\tilde\phi_{i,k}
&=
\operatorname{Norm}_{j\in\mathcal{A}_{x,k}}
\left[
\phi_{j,k}
\right]_i,\\
A^{\mathrm{TCPO}}_{i,k}
&=
\gamma^{k-1}
\tilde\phi_{i,k}.
\end{aligned}
\]
Here $[\cdot]_i$ denotes the normalized value corresponding to trajectory $i$. When $|\mathcal{A}_{x,k}|<2$, normalization is skipped and $\gamma^{k-1}\phi_{i,k}$ is used directly.

The turn-level advantage is broadcast to all assistant tokens in the corresponding turn:
\[
A_{i,k,t}=A^{\mathrm{TCPO}}_{i,k},
\qquad
a_{i,k,t}\in y_{i,k}.
\]
TCPO replaces only the advantage estimator in a GRPO-style clipped objective:
\[
\begin{aligned}
\mathcal{L}_{\mathrm{TCPO}}(\theta)
&=
-\mathbb{E}_{i,k,t}
\left[
\min(\ell_{i,k,t},\bar\ell_{i,k,t})
\right],\\
\ell_{i,k,t}
&=
\rho_{i,k,t}(\theta)A_{i,k,t},\\
\bar\ell_{i,k,t}
&=
\bar\rho_{i,k,t}(\theta)A_{i,k,t},\\
\bar\rho_{i,k,t}(\theta)
&=
\operatorname{clip}(\rho_{i,k,t}(\theta),1-\epsilon,1+\epsilon).
\end{aligned}
\]
The likelihood ratio is
\[
\rho_{i,k,t}(\theta)
=
\frac{
\pi_{\theta}(a_{i,k,t}\mid h_{i,k},a_{i,k,<t})
}{
\pi_{\mathrm{old}}(a_{i,k,t}\mid h_{i,k},a_{i,k,<t})
}.
\]

\begin{table*}[!t]
\centering
\caption{Main results on math and code tasks. Baselines include GRPO~\citep{shao2024deepseekmath}, MT-GRPO~\citep{zeng2025reinforcing}, and GVPO~\citep{dai2026group}. Pass@8 is best-turn Pass@8, and Avg. Turns is computed over successful trajectories.}
\label{tab:main_results}
\resizebox{0.98\linewidth}{!}{%
\begin{tabular}{lcccccccc}
\toprule
\textbf{Method} & \multicolumn{4}{c}{\textbf{Math}} & \multicolumn{4}{c}{\textbf{Code}} \\
\cmidrule(lr){2-5}\cmidrule(lr){6-9}
 & \multicolumn{2}{c}{\textbf{MATH-500}} & \multicolumn{2}{c}{\textbf{AIME}} & \multicolumn{2}{c}{\textbf{LiveCodeBench}} & \multicolumn{2}{c}{\textbf{HumanEval}} \\
\cmidrule(lr){2-3}\cmidrule(lr){4-5}\cmidrule(lr){6-7}\cmidrule(lr){8-9}
 & \textbf{Pass@8} & \textbf{Avg. Turns} & \textbf{Pass@8} & \textbf{Avg. Turns} & \textbf{Pass@8} & \textbf{Avg. Turns} & \textbf{Pass@8} & \textbf{Avg. Turns} \\
\midrule
\multicolumn{9}{c}{\textit{Qwen3-4B}} \\
\cmidrule(lr){1-9}
Base Model & 79.6\% & 1.42 & 47.8\% & 1.53 & 38.7\% & 1.96 & 84.1\% & 1.35 \\
GRPO & 80.8\% & 1.26 & 50.0\% & 1.49 & 42.2\% & 1.61 & 85.9\% & 1.31 \\
MT-GRPO & 82.4\% & 1.21 & 48.9\% & 1.37 & 43.5\% & 1.42 & 86.6\% & 1.23 \\
GVPO & 85.2\% & 1.22 & 51.1\% & 1.36 & 44.3\% & 1.39 & 86.6\% & 1.18 \\
\rowcolor{gray!10}TCPO & \textbf{86.8\%} & \textbf{1.19} & \textbf{51.1\%} & \textbf{1.33} & \textbf{45.1\%} & \textbf{1.27} & \textbf{87.2\%} & \textbf{1.12} \\
\cmidrule(lr){1-9}
\multicolumn{9}{c}{\textit{DeepSeek-R1-Distill-Llama-8B}} \\
\cmidrule(lr){1-9}
Base Model & 29.8\% & 2.25 & 21.1\% & 2.67 & 12.6\% & 2.83 & 84.8\% & 1.33 \\
GRPO & 56.6\% & 1.84 & 35.6\% & 2.08 & 27.9\% & 2.45 & 87.2\% & 1.24 \\
MT-GRPO & 59.0\% & 1.71 & 35.6\% & 1.92 & 30.3\% & 2.21 & 87.2\% & 1.19 \\
GVPO & 62.8\% & 1.62 & 36.7\% & 1.87 & 31.8\% & 2.13 & 87.8\% & 1.19 \\
\rowcolor{gray!10}TCPO & \textbf{65.2\%} & \textbf{1.59} & \textbf{38.9\%} & \textbf{1.86} & \textbf{32.2\%} & \textbf{2.13} & \textbf{88.4\%} & \textbf{1.15} \\
\bottomrule
\end{tabular}%
}
\end{table*}

\begin{table*}[!t]
\centering
\small
\begin{minipage}[t]{0.52\linewidth}
\centering
\resizebox{\linewidth}{!}{%
\begin{tabular}{lcccccc}
\toprule
\textbf{Method}
& \multicolumn{2}{c}{\textbf{Dev}}
& \multicolumn{2}{c}{\textbf{Test-N}}
& \multicolumn{2}{c}{\textbf{Test-C}} \\
\cmidrule(lr){2-3}
\cmidrule(lr){4-5}
\cmidrule(lr){6-7}
& \textbf{TGC} & \textbf{SGC}
& \textbf{TGC} & \textbf{SGC}
& \textbf{TGC} & \textbf{SGC} \\
\midrule
\multicolumn{7}{c}{\textit{Qwen2.5-32B-Instruct}} \\
\cmidrule(lr){1-7}
GVPO
& 84.2 & 73.7
& 72.6 & 55.4
& 49.4 & 28.8 \\
\rowcolor{gray!10}
TCPO
& \textbf{88.3} & \textbf{75.0}
& \textbf{74.4} & \textbf{57.1}
& \textbf{49.9} & \textbf{28.8} \\
\bottomrule
\end{tabular}%
}
\caption{Comparison on the AppWorld benchmark with GVPO~\citep{dai2026group}, following the original GVPO AppWorld setup with Qwen2.5-32B-Instruct. TGC and SGC denote task-level and scenario-level goal completion on the AppWorld~\citep{trivedi2024appworld} Dev, Test-N, and Test-C splits.}
\label{tab:qwen25_32b_gvpo_tcpo}
\end{minipage}\hfill
\begin{minipage}[t]{0.44\linewidth}
\centering
\resizebox{\linewidth}{!}{%
\begin{tabular}{lcc}
\toprule
\textbf{Variant} & \textbf{MATH-500} & \textbf{LiveCodeBench} \\
\midrule
Traj. reward & 56.6\% & 27.9\% \\
Turn reward & 59.0\% & 30.3\% \\
$\Delta$ reward & 59.8\% & 30.4\% \\
Retro & 62.2\% & 31.5\% \\
Retro + Hin. & 63.4\% & 31.7\% \\
\rowcolor{gray!10}TCPO & \textbf{65.2\%} & \textbf{32.2\%} \\
\bottomrule
\end{tabular}%
}
\caption{Credit-signal ablation with DeepSeek-R1-Distill-Llama-8B. $\Delta$ reward uses adjacent verifier-score differences. Retro combines improvement, keep, and regression terms. Hin. denotes hindsight delayed credit.}
\label{tab:ablation}
\end{minipage}
\end{table*}

\section{Experiments}

This section organizes the experiments around three questions: (i) does TCPO improve or match strong baselines across model scales, task domains, and verifier types; (ii) do the gains come from more informative reference-aware credit construction rather than denser supervision alone; and (iii) can selective fixed-history counterfactual estimation improve performance with modest training overhead?

\subsection{Experimental Setup}

\paragraph{Tasks and evaluation.}
We evaluate TCPO on mathematical reasoning, code generation, and interactive agent tasks. For mathematics, we use MATH-500~\citep{hendrycks2021math} and AIME 2024--2026~\citep{dekoninck2026matharena}; for code, we use LiveCodeBench v6~\citep{jain2024livecodebench} and HumanEval~\citep{chen2021codex}; for agents, we use AppWorld~\citep{trivedi2024appworld}. In the math and code settings, the model generates a candidate answer at each turn, and a verifier returns a turn-level score. During training, trajectories always run for the fixed $K$-turn budget, including turns after the first success, so that preservation and regression remain observable. During evaluation, success is recorded at the first correct turn, and generation may terminate immediately thereafter. For math tasks, success means producing the correct final answer; for code tasks, success means passing all test cases. AppWorld uses its native state-based verifier and task-completion metrics, which makes it a representative interactive agent setting.

\begin{table*}[!t]
\centering
\small
\caption{Counterfactual budgeting with DeepSeek-R1-Distill-Llama-8B. TCPO is the default budgeted setting used in the main experiments. Rand.-$L$ matches the per-prompt Top-$L$ budget with random eligible turns; Top-$L$ selects high-surprisal eligible turns per prompt. CF-Aug. additionally trains on selected counterfactual samples. Train overhead is end-to-end wall-clock time relative to No-CF under identical hardware and training steps.}
\label{tab:cf_budget}
\resizebox{0.88\linewidth}{!}{%
\begin{tabular}{lllcccc}
\toprule
\textbf{Variant} & \textbf{CF Selection} & \textbf{CF Usage} &
\multicolumn{2}{c}{\textbf{MATH-500}} &
\multicolumn{2}{c}{\textbf{LiveCodeBench}} \\
\cmidrule(lr){4-5}\cmidrule(lr){6-7}
& & &
\textbf{Pass@8} & \textbf{Train Overhead} &
\textbf{Pass@8} & \textbf{Train Overhead} \\
\midrule
No-CF & -- & -- & 63.4\% & 1.00$\times$ & 31.7\% & 1.00$\times$ \\
Random CF & Rand.-$L$ & Credit & 63.4\% & 1.03$\times$ & 31.8\% & 1.05$\times$ \\
\rowcolor{gray!10}TCPO & Top-$L$ & Credit & 65.2\% & 1.03$\times$ & 32.2\% & 1.05$\times$ \\
All CF & All & Credit & 66.8\% & 1.11$\times$ & 33.1\% & 1.16$\times$ \\
CF-Aug. & Top-$L$ & Credit+Train & 69.2\% & 1.25$\times$ & 34.5\% & 1.33$\times$ \\
\bottomrule
\end{tabular}%
}
\end{table*}

\paragraph{Models and training data.}
For math and code, we train on DAPO-Math-17K~\citep{yu2025dapo} and TACO-Verified~\citep{likaixin2024taco-verified}, respectively. HumanEval is used only for evaluation. We conduct the main experiments on Qwen3-4B and DeepSeek-R1-Distill-Llama-8B. Training for these tasks is implemented in verl~\citep{sheng2024hybridflow} with vLLM-based asynchronous rollouts and FSDP2 for the actor and reference models, and runs on $4$ NVIDIA H200 GPUs. We use global batch sizes of 64 for math and 32 for code, a learning rate of $1\times10^{-6}$, and train for one epoch. For AppWorld, we use Qwen2.5-32B-Instruct and follow the original GVPO AppWorld setup~\citep{dai2026group}, including the same training split, rollout budget, decoding settings, and evaluation protocol. This allows us to test whether TCPO is effective across model scales, task domains, and verifier types, rather than being tied to a particular benchmark or architecture.

\paragraph{Rollout settings.}
For math and code, we sample $G=8$ multi-turn trajectories per prompt. We set the maximum number of turns to $K=3$ for math and $K=5$ for code. The maximum response length is 8,192 tokens for math and 16,384 tokens for code. AppWorld follows the original GVPO rollout budget. For a given initial model, all methods use the same prompts, verifier, base rollout configuration, generation settings, and token-level objective. TCPO additionally applies budgeted branching to selected high-surprisal turns. For each prompt, TCPO selects up to $L=5$ eligible turns for math and $L=10$ for code and AppWorld, and samples $M=8$ fixed-history alternatives for each selected turn. Full optimization and sampling settings are provided in the supplementary material.

\paragraph{Baselines.}
We compare against Base, GRPO, MT-GRPO, GVPO, and several credit-assignment ablations. Base is the initial model before multi-turn RL. Base-model results use the same multi-turn repair prompt, generation limits, verifier, and best-turn protocol as the RL methods, and are therefore in-protocol baselines rather than official single-turn benchmark results. GRPO optimizes with trajectory-level outcome reward under a group-relative objective. MT-GRPO introduces turn-level reward allocation in multi-turn trajectories to alleviate the coarse granularity of trajectory-level reward. GVPO uses process-verifiable signals to build a finer-grained optimization target. For math and code, we adapt GVPO by applying its process-verifiable success/failure functions at the turn level. In the ablation study, raw turn score uses the current verifier score as the credit for that turn; adjacent score difference uses the difference between the current score and the previous score, i.e., $r_{i,k}-r_{i,k-1}$, as the current turn credit; the remaining variants progressively add retrospective credit, hindsight delayed credit, and selective counterfactual estimation. Unless otherwise specified, the variants in Tables~\ref{tab:ablation} and~\ref{tab:cf_budget} use the same turn-level normalization and early-turn weighting. All methods share the same base rollout protocol and verifier feedback. TCPO differs by converting these base traces into reference-aware turn credit and by adding budgeted post-rollout counterfactual branches for selected turns.

\paragraph{Metrics.}
For math and code tasks, we report best-turn Pass@8: a prompt is counted as successful if any of the $8$ sampled trajectories reaches a correct turn. This metric matches the verifier-guided multi-turn inference setting, where the model may revise its answer multiple times and stop as soon as one turn becomes correct. We also report Avg. Turns to Success, computed over successful trajectories only, to measure how quickly the model reaches a correct state. For AppWorld, we report TGC and SGC, which measure task-level and scenario-level goal completion, respectively.

\paragraph{Performance across model scales and task domains.}
Table~\ref{tab:main_results} summarizes the main results on math and code tasks. Across both Qwen3-4B and DeepSeek-R1-Distill-Llama-8B, TCPO achieves the best or tied-best best-turn Pass@8 on every benchmark and generally reduces the average turns needed to reach success. On Qwen3-4B, it improves over MT-GRPO by $+4.4$, $+2.2$, $+1.6$, and $+0.6$ points on MATH-500, AIME, LiveCodeBench, and HumanEval, respectively. It also improves over GVPO on MATH-500, LiveCodeBench, and HumanEval, while matching it on AIME. These results suggest that even when turn-level verifier feedback is available, the reference used to convert scores into credit remains important.

The gains are larger on DeepSeek-R1-Distill-Llama-8B, where TCPO improves over MT-GRPO by $+6.2$, $+3.3$, $+1.9$, and $+1.2$ points and outperforms GVPO on all four tasks. This weaker initialization produces more incorrect attempts, partial repairs, and post-success regressions during multi-turn rollouts, making the raw score trace harder to interpret. The stronger gains in this setting are consistent with TCPO's design: retrospective, hindsight, and counterfactual references help distinguish useful repair behavior from redundant or harmful turns, rather than assigning credit only from the final trajectory outcome or adjacent score changes.

Table~\ref{tab:qwen25_32b_gvpo_tcpo} evaluates TCPO on AppWorld under the same setting as the original GVPO study. TCPO improves both TGC and SGC on Dev and Test-N, and improves TGC on Test-C while matching SGC. Unlike the math and code environments, AppWorld requires longer stateful interactions, so later success depends on preserving and updating task state across many tool-use steps. The AppWorld results therefore indicate that the score-to-credit conversion problem is not specific to symbolic reasoning or program repair, and that TCPO can transfer to interactive agent tasks with different verifier structure.


\begin{figure*}[t]
\centering
\begin{minipage}[t]{0.32\textwidth}
\centering
\includegraphics[width=\linewidth]{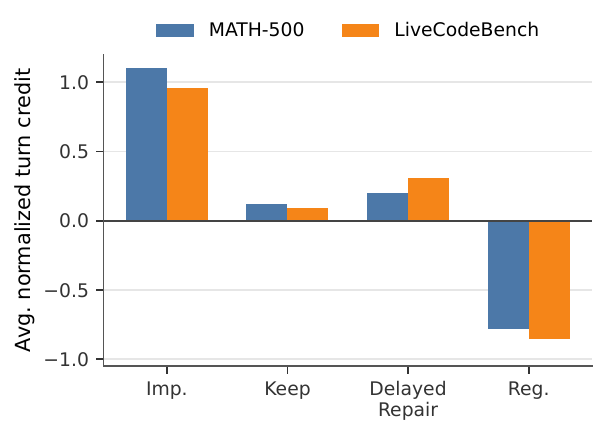}
{\small (a) Credit by turn type}
\end{minipage}
\hfill
\begin{minipage}[t]{0.32\textwidth}
\centering
\includegraphics[width=\linewidth]{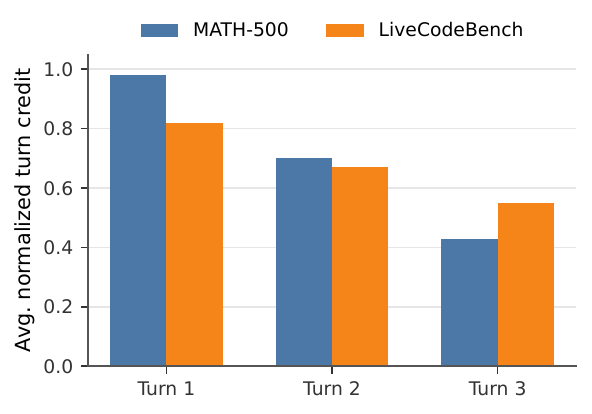}
{\small (b) Credit by turn index}
\end{minipage}
\hfill
\begin{minipage}[t]{0.32\textwidth}
\centering
\includegraphics[width=\linewidth]{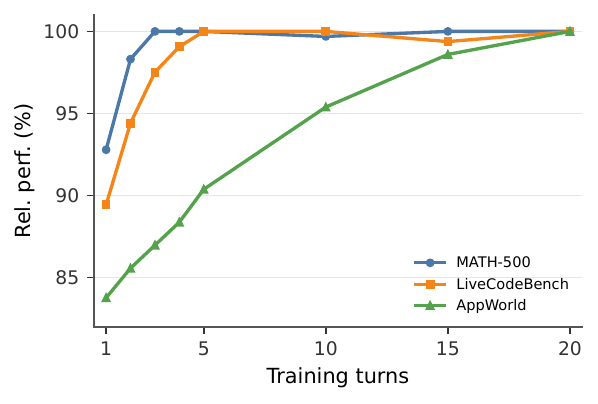}
{\small (c) Number of training turns}
\end{minipage}
\caption{Analysis diagnostics. (a) Average normalized credit assigned to different turn types, where Imp. denotes improvement and Reg. denotes regression. (b) Average normalized credit by turn index on successful repairs. (c) Sensitivity to the number of training turns; performance is normalized by the saturated score of each dataset, using Pass@8 for MATH-500 and LiveCodeBench and TGC for AppWorld.}
\label{fig:analysis_diagnostics}
\end{figure*}

\paragraph{Where Do the Gains Come From?}

Table~\ref{tab:ablation} isolates the effect of each credit reference. Trajectory-level reward is the coarsest signal because it assigns the same supervision to failed attempts, effective repairs, and post-success regressions. Raw turn score is denser, but it evaluates the current output rather than the turn's marginal contribution. Adjacent score difference captures local change, but only relative to the immediately preceding score. Retrospective credit improves over these local baselines, increasing DeepSeek MATH-500 from 59.8 to 62.2 and LiveCodeBench from 30.4 to 31.5 over adjacent score difference, supporting the best-so-far score as a more informative reference.

Retrospective credit still cannot resolve pre-success non-improving turns whose value appears only later. Adding hindsight delayed credit raises performance to 63.4 on MATH-500 and 31.7 on LiveCodeBench, suggesting that realized future-best traces provide useful delayed signals beyond local score differences. Full TCPO further reaches 65.2 and 32.2 by applying selective fixed-history counterfactual estimation to high-surprisal turns. Since these comparisons fix the same history and sample alternative outputs, they provide a more context-matched estimate than hindsight alone. Overall, the ablation supports the central claim that multi-turn verifier scores become useful when converted through appropriate references, not merely when made denser.

\paragraph{Efficiency of Counterfactual Estimation}

Table~\ref{tab:cf_budget} studies the trade-off between counterfactual gain and training overhead. The retrospective+hindsight variant already improves over the baseline without counterfactual estimation, but it still depends on indirect references from existing rollouts. Randomly selecting turns for counterfactual estimation brings almost no additional gain, which indicates that extra counterfactual computation alone is not sufficient; the key is to spend it on the turns where credit is most uncertain.

TCPO selects high-surprisal eligible turns and applies fixed-history counterfactual estimation only to them. With only about $1.03\times$ to $1.05\times$ training overhead, TCPO improves MATH-500 from 63.4 to 65.2 and LiveCodeBench from 31.7 to 32.2. By contrast, applying counterfactual estimation to all candidate turns yields somewhat higher performance, but the training overhead increases to $1.11\times$ to $1.16\times$. Extending the branch candidates into the training batch achieves the strongest numbers, but the cost rises to $1.25\times$ to $1.33\times$, which is beyond the default setting of this paper.

These results show that TCPO's counterfactual module is not designed to maximize raw sampling volume. Instead, it allocates a small counterfactual budget to high-surprisal turns, which serve as a proxy for credit ambiguity. The matched-budget comparison shows that this targeted selection is more effective than random branching. Top-$L$ selection offers a favorable accuracy--cost trade-off while keeping the cost close to that of the original rollout procedure. In this sense, TCPO improves over pure hindsight estimation while avoiding the higher cost of full branching or branch augmentation.

\subsection{Credit Dynamics and Training Horizon}

Figure~\ref{fig:analysis_diagnostics} shows that TCPO assigns positive credit to improvement, preservation, and delayed repair, while penalizing post-success regression. This pattern matches the intended repair dynamics: an update should be rewarded when it improves or preserves a verified state, penalized when it breaks a solved state, and still be creditable when it enables later repair. Earlier turns also receive larger normalized credit, consistent with the early-turn prior and the goal of encouraging useful repairs earlier in the trajectory.

The number-of-training-turns analysis shows that useful turn credit differs across tasks. Relative performance denotes the score achieved by a model trained with a given number of turns, normalized by the best score attained for that task across the tested turn budgets. Under this view, MATH-500 approaches saturation by three turns, LiveCodeBench reaches its plateau around five turns, and AppWorld rises from about 84\% at one turn to 100\% at twenty turns. As trajectories grow longer, trajectory-level reward increasingly mixes early failures, mid-course repairs, and late redundant actions. Together with the ablations and counterfactual budget study, these results indicate that TCPO's gains come from reference-aware credit construction under the same base rollout budget, with selective counterfactual refinement adding only modest training overhead.

\section{Conclusion}

This paper presents TCPO, a turn-level policy optimization method for verifier-guided multi-turn reinforcement learning. TCPO reframes per-turn verifier scores as observations that must be converted into credit through suitable retrospective, hindsight, and fixed-history counterfactual references. These reference-aware signals target turns that repair, preserve, or degrade a solution. Experiments across mathematical reasoning, code generation, and long-horizon agentic tasks show that this reference-aware credit improves final success rates and often reaches verified solutions in fewer turns, while the counterfactual component remains computationally controlled through high-surprisal selection.

\newpage
\bibliographystyle{plainnat}
\bibliography{tcpo}

@misc{chen2021codex,
  title = {Evaluating Large Language Models Trained on Code},
  author = {Chen, Mark and Tworek, Jerry and Jun, Heewoo and Yuan, Qiming and Pinto, Henrique Ponde de Oliveira and Kaplan, Jared and Edwards, Harri and Burda, Yuri and Joseph, Nicholas and Brockman, Greg and Ray, Alex and Puri, Raul and Krueger, Gretchen and Petrov, Michael and Khlaaf, Heidy and Sastry, Girish and Mishkin, Pamela and Chan, Brooke and Gray, Scott and Ryder, Nick and Pavlov, Mikhail and Power, Alethea and Kaiser, Lukasz and Bavarian, Mohammad and Winter, Clemens and Tillet, Philippe and Such, Felipe Petroski and Cummings, Dave and Plappert, Matthias and Chantzis, Fotios and Barnes, Elizabeth and Herbert-Voss, Ariel and Guss, William Hebgen and Nichol, Alex and Paino, Alex and Tezak, Nikolas and Tang, Jie and Babuschkin, Igor and Balaji, Suchir and Jain, Shantanu and Saunders, William and Hesse, Christopher and Carr, Andrew N. and Leike, Jan and Achiam, Josh and Misra, Vedant and Morikawa, Evan and Radford, Alec and Knight, Matthew and Brundage, Miles and Murati, Mira and Mayer, Katie and Welinder, Peter and McGrew, Bob and Amodei, Dario and McCandlish, Sam and Sutskever, Ilya and Zaremba, Wojciech},
  year = {2021},
  eprint = {2107.03374},
  archivePrefix = {arXiv},
  primaryClass = {cs.LG},
}

@misc{shao2024deepseekmath,
  title = {DeepSeekMath: Pushing the Limits of Mathematical Reasoning in Open Language Models},
  author = {Shao, Zhihong and Wang, Peiyi and Zhu, Qihao and Xu, Runxin and Song, Junxiao and Bi, Xiao and Zhang, Haowei and Zhang, Mingchuan and Li, Y. K. and Wu, Y. and Guo, Daya},
  year = {2024},
  eprint = {2402.03300},
  archivePrefix = {arXiv},
  primaryClass = {cs.CL},
}

@inproceedings{lightman2024let,
  title={Let's verify step by step},
  author={Lightman, Hunter and Kosaraju, Vineet and Burda, Yuri and Edwards, Harrison and Baker, Bowen and Lee, Teddy and Leike, Jan and Schulman, John and Sutskever, Ilya and Cobbe, Karl},
  booktitle={International Conference on Learning Representations},
  volume={2024},
  pages={39578--39601},
  year={2024}
}

@misc{hendrycks2021math,
  title = {Measuring Mathematical Problem Solving With the {MATH} Dataset},
  author = {Hendrycks, Dan and Burns, Collin and Kadavath, Saurav and Arora, Akul and Basart, Steven and Tang, Eric and Song, Dawn and Steinhardt, Jacob},
  year = {2021},
  eprint = {2103.03874},
  archivePrefix = {arXiv},
  primaryClass = {cs.LG},
}

@misc{jain2024livecodebench,
  title = {{LiveCodeBench}: Holistic and Contamination Free Evaluation of Large Language Models for Code},
  author = {Jain, Naman and Han, King and Gu, Alex and Li, Wen-Ding and Yan, Fanjia and Zhang, Tianjun and Wang, Sida and Solar-Lezama, Armando and Sen, Koushik and Stoica, Ion},
  year = {2024},
  eprint = {2403.07974},
  archivePrefix = {arXiv},
  primaryClass = {cs.SE},
}

@misc{yu2025dapo,
  title = {{DAPO}: An Open-Source {LLM} Reinforcement Learning System at Scale},
  author = {Yu, Qiying and Zhang, Zheng and Zhu, Ruofei and Yuan, Yufeng and Zuo, Xiaochen and Yue, Yu and Dai, Weinan and Fan, Tiantian and Liu, Gaohong and Liu, Lingjun and Liu, Xin and Lin, Haibin and Lin, Zhiqi and Ma, Bole and Sheng, Guangming and Tong, Yuxuan and Zhang, Chi and Zhang, Mofan and Zhang, Wang and Zhu, Hang and Zhu, Jinhua and Chen, Jiaze and Chen, Jiangjie and Wang, Chengyi and Yu, Hongli and Song, Yuxuan and Wei, Xiangpeng and Zhou, Hao and Liu, Jingjing and Ma, Wei-Ying and Zhang, Ya-Qin and Yan, Lin and Qiao, Mu and Wu, Yonghui and Wang, Mingxuan},
  year = {2025},
  eprint = {2503.14476},
  archivePrefix = {arXiv},
  primaryClass = {cs.LG},
}

@misc{likaixin2024taco-verified,
  author       = {Kaixin Li},
  title        = {Verified TACO Problems},
  year         = {2024},
  url          = {https://huggingface.co/datasets/likaixin/TACO-verified},
  howpublished = {\url{https://huggingface.co/datasets/likaixin/TACO-verified}},
}

@article{uesato2022solving,
  title={Solving math word problems with process-and outcome-based feedback},
  author={Uesato, Jonathan and Kushman, Nate and Kumar, Ramana and Song, Francis and Siegel, Noah and Wang, Lisa and Creswell, Antonia and Irving, Geoffrey and Higgins, Irina},
  journal={arXiv preprint arXiv:2211.14275},
  year={2022}
}

@inproceedings{wang2024math,
  title={Math-shepherd: Verify and reinforce llms step-by-step without human annotations},
  author={Wang, Peiyi and Li, Lei and Shao, Zhihong and Xu, Runxin and Dai, Damai and Li, Yifei and Chen, Deli and Wu, Yu and Sui, Zhifang},
  booktitle={Proceedings of the 62nd Annual Meeting of the Association for Computational Linguistics (Volume 1: Long Papers)},
  pages={9426--9439},
  year={2024}
}

@article{zhou2025sweet,
  title={Sweet-rl: Training multi-turn llm agents on collaborative reasoning tasks},
  author={Zhou, Yifei and Jiang, Song and Tian, Yuandong and Weston, Jason and Levine, Sergey and Sukhbaatar, Sainbayar and Li, Xian},
  journal={arXiv preprint arXiv:2503.15478},
  year={2025}
}

@inproceedings{chen2025context,
  title={Context-lite multi-turn reinforcement learning for LLM agents},
  author={Chen, Wentse and Chen, Jiayu and Zhu, Hao and Schneider, Jeff},
  booktitle={ES-FoMo III: 3rd Workshop on Efficient Systems for Foundation Models},
  year={2025}
}

@inproceedings{zeng2025reinforcing,
  title={Reinforcing multi-turn reasoning in llm agents via turn-level credit assignment},
  author={Zeng, Siliang and Wei, Quan and Brown, William and Frunza, Oana and Nevmyvaka, Yuriy and Zhao, Yang Katie and Hong, Mingyi},
  booktitle={ICML 2025 Workshop on Computer Use Agents},
  year={2025}
}

@inproceedings{dai2026group,
  title={Group Verification-based Policy Optimization for Interactive Coding Agents},
  author={Dai, Silong and Sun, Changzhi and Wu, Haolun and Zheng, Huanran and Ji, Tao and Yan, Junchi and Wu, Yuanbin and Zhang, Dell and Wang, Xiaoling and Li, Xuelong},
  booktitle={The Fourteenth International Conference on Learning Representations},
  year={2026}
}

@inproceedings{trivedi2024appworld,
  title={Appworld: A controllable world of apps and people for benchmarking interactive coding agents},
  author={Trivedi, Harsh and Khot, Tushar and Hartmann, Mareike and Manku, Ruskin and Dong, Vinty and Li, Edward and Gupta, Shashank and Sabharwal, Ashish and Balasubramanian, Niranjan},
  booktitle={Proceedings of the 62nd Annual Meeting of the Association for Computational Linguistics (Volume 1: Long Papers)},
  pages={16022--16076},
  year={2024}
}

@inproceedings{li2026turn,
  title={Turn-ppo: Turn-level advantage estimation with ppo for improved multi-turn rl in agentic llms},
  author={Li, Junbo and Zhou, Peng and Meng, Rui and Vadera, Meet P and Li, Lihong and Li, Yang},
  booktitle={Findings of the Association for Computational Linguistics: EACL 2026},
  pages={6227--6243},
  year={2026}
}

@article{tao2026trace,
  title={TRACE: Turn-level Reward Assignment via Credit Estimation for Long-Horizon Agents},
  author={Tao, Leitian and Peng, Baolin and Yao, Wenlin and Ge, Tao and Cheng, Hao and Wang, Mike Hang and Gao, Jianfeng and Li, Sharon},
  journal={arXiv preprint arXiv:2607.13988},
  year={2026}
}

@article{chen2026learning,
  title={Learning to reason with search for llms via reinforcement learning},
  author={Chen, Mingyang and Sun, Linzhuang and Li, Tianpeng and Sun, Haoze and Zhu, Chenzheng and Wang, Haofen and Pan, Jeff and Zhang, Wen and Chen, Huajun and Yang, Fan and others},
  journal={Advances in Neural Information Processing Systems},
  volume={38},
  pages={85287--85307},
  year={2026}
}

@article{jin2025search,
  title={Search-r1: Training llms to reason and leverage search engines with reinforcement learning},
  author={Jin, Bowen and Zeng, Hansi and Yue, Zhenrui and Yoon, Jinsung and Arik, Sercan and Wang, Dong and Zamani, Hamed and Han, Jiawei},
  journal={arXiv preprint arXiv:2503.09516},
  year={2025}
}

@article{feng2025retool,
  title={Retool: Reinforcement learning for strategic tool use in llms},
  author={Feng, Jiazhan and Huang, Shijue and Qu, Xingwei and Zhang, Ge and Qin, Yujia and Zhong, Baoquan and Jiang, Chengquan and Chi, Jinxin and Zhong, Wanjun},
  journal={arXiv preprint arXiv:2504.11536},
  year={2025}
}

@article{li2025torl,
  title={Torl: Scaling tool-integrated rl},
  author={Li, Xuefeng and Zou, Haoyang and Liu, Pengfei},
  journal={arXiv preprint arXiv:2503.23383},
  year={2025}
}

@article{wang2025otc,
  title={Otc: Optimal tool calls via reinforcement learning},
  author={Wang, Hongru and Qian, Cheng and Zhong, Wanjun and Chen, Xiusi and Qiu, Jiahao and Huang, Shijue and Jin, Bowen and Wang, Mengdi and Wong, Kam-Fai and Ji, Heng},
  journal={arXiv e-prints},
  pages={arXiv--2504},
  year={2025}
}

@article{zhang2025nemotron,
  title={Nemotron-research-tool-n1: Exploring tool-using language models with reinforced reasoning},
  author={Zhang, Shaokun and Dong, Yi and Zhang, Jieyu and Kautz, Jan and Catanzaro, Bryan and Tao, Andrew and Wu, Qingyun and Yu, Zhiding and Liu, Guilin},
  journal={arXiv preprint arXiv:2505.00024},
  year={2025}
}

@article{singh2025agentic,
  title={Agentic Reasoning and Tool Integration for LLMs via Reinforcement Learning},
  author={Singh, Joykirat and Magazine, Raghav and Pandya, Yash and Nambi, Akshay},
  journal={arXiv preprint arXiv:2505.01441},
  year={2025}
}

@inproceedings{yao2020keep,
  title={Keep calm and explore: Language models for action generation in text-based games},
  author={Yao, Shunyu and Rao, Rohan and Hausknecht, Matthew and Narasimhan, Karthik},
  booktitle={Proceedings of the 2020 Conference on Empirical Methods in Natural Language Processing (EMNLP)},
  pages={8736--8754},
  year={2020}
}

@inproceedings{carta2023grounding,
  title={Grounding large language models in interactive environments with online reinforcement learning},
  author={Carta, Thomas and Romac, Cl{\'e}ment and Wolf, Thomas and Lamprier, Sylvain and Sigaud, Olivier and Oudeyer, Pierre-Yves},
  booktitle={International conference on machine learning},
  pages={3676--3713},
  year={2023},
  organization={PMLR}
}

@article{zhai2024fine,
  title={Fine-tuning large vision-language models as decision-making agents via reinforcement learning},
  author={Zhai, Yuexiang and Bai, Hao and Lin, Zipeng and Pan, Jiayi and Tong, Shengbang and Zhou, Yifei and Suhr, Alane and Xie, Saining and LeCun, Yann and Ma, Yi and others},
  journal={Advances in neural information processing systems},
  volume={37},
  pages={110935--110971},
  year={2024}
}

@article{wang2025ragen,
  title={Ragen: Understanding self-evolution in llm agents via multi-turn reinforcement learning},
  author={Wang, Zihan and Wang, Kangrui and Wang, Qineng and Zhang, Pingyue and Li, Linjie and Yang, Zhengyuan and Jin, Xing and Yu, Kefan and Nguyen, Minh Nhat and Liu, Licheng and others},
  journal={arXiv preprint arXiv:2504.20073},
  year={2025}
}

@article{yao2022webshop,
  title={Webshop: Towards scalable real-world web interaction with grounded language agents},
  author={Yao, Shunyu and Chen, Howard and Yang, John and Narasimhan, Karthik},
  journal={Advances in Neural Information Processing Systems},
  volume={35},
  pages={20744--20757},
  year={2022}
}

@article{chen2025reinforcement,
  title={Reinforcement Learning for Long-Horizon Interactive LLM Agents},
  author={Chen, Kevin and Cusumano-Towner, Marco and Huval, Brody and Petrenko, Aleksei and Hamburger, Jackson and Koltun, Vladlen and Kr{\"a}henb{\"u}hl, Philipp},
  journal={arXiv preprint arXiv:2502.01600},
  year={2025}
}

@article{bai2024digirl,
  title={Digirl: Training in-the-wild device-control agents with autonomous reinforcement learning},
  author={Bai, Hao and Zhou, Yifei and Pan, Jiayi and Cemri, Mert and Suhr, Alane and Levine, Sergey and Kumar, Aviral},
  journal={Advances in Neural Information Processing Systems},
  volume={37},
  pages={12461--12495},
  year={2024}
}

@article{jin2025empirical,
  title={An Empirical Study on Reinforcement Learning for Reasoning-Search Interleaved LLM Agents},
  author={Jin, Bowen and Yoon, Jinsung and Kargupta, Priyanka and Arik, Sercan O and Han, Jiawei},
  journal={arXiv preprint arXiv:2505.15117},
  year={2025}
}

@inproceedings{tran2026exploiting,
  title={Exploiting Tree Structure for Credit Assignment in Reinforcement Learning with Large Language Models},
  author={Tran, Hieu and Yao, Zonghai and Yu, Hong},
  booktitle={Findings of the Association for Computational Linguistics: ACL 2026},
  pages={10795--10810},
  year={2026}
}

@article{zong20262,
  title={A$T^{2}$PO: Agentic Turn-based Policy Optimization via Tree Search},
  author={Zong, Zefang and Chen, Dingwei and Li, Yang and Yi, Qi and Zhou, Bo and Li, Chengming and Qian, Bo and Chen, Peng and Jiang, Jie},
  journal={arXiv preprint arXiv:2601.04767},
  year={2026}
}

@article{dekoninck2026matharena,
  title={Beyond Benchmarks: MathArena as an Evaluation Platform for Mathematics with LLMs},
  author={Dekoninck, Jasper and Jovanovi{\'c}, Nikola and Gehrunger, Tim and R{\"o}gnvaldsson, K{\'a}ri and Petrov, Ivo and Sun, Chenhao and Vechev, Martin},
  year={2026},
  eprint={2605.00674},
  archivePrefix={arXiv},
  primaryClass={cs.CL},
  url={https://arxiv.org/abs/2605.00674}
}

@article{sheng2024hybridflow,
  title={HybridFlow: A Flexible and Efficient RLHF Framework},
  author={Sheng, Guangming and Zhang, Chi and Ye, Zilingfeng and Wu, Xibin and Zhang, Wang and Zhang, Ru and Peng, Yanghua and Lin, Haibin and Wu, Chuan},
  journal={arXiv preprint arXiv:2409.19256},
  year={2024}
}

\end{document}